\RequirePackage{fix-cm}
\RequirePackage{rotating}
\documentclass[twocolumn]{svjour3}          
\usepackage{natbib}
\smartqed  
\usepackage{graphicx}
\usepackage[table,xcdraw]{xcolor}
\usepackage{lscape}
\usepackage{adjustbox}
\usepackage[hyphens]{url}
\urldef{\footcamurl}\url{http://mi.eng.cam.ac.uk/research/projects/VideoRec/CamVid/}
\usepackage{hyperref}
\journalname{}
\begin{document}

\title{Road surface detection and differentiation considering surface damages \thanks{This study was financed in part by the Coordenação de Aperfeiçoamento de Pessoal de Nível Superior - Brasil (CAPES) - Finance Code 001. CAPES (Brazilian Federal Agency for Support and Evaluation of Graduate Education). It was also supported by the Brazilian National Institute for Digital Convergence (INCoD), a research unit of the Brazilian National Institutes for Science and Technology Program (INCT) of the Brazilian National Council for Science and Technology (CNPq).}
}


\author{Thiago Rateke\textsuperscript{1,2}         \and
        Aldo von Wangenheim\textsuperscript{1,2} 
}


\institute{Thiago Rateke \at
              \email{thiago@incod.ufsc.br}           
           \and
           Aldo von Wangenheim \at
              \email{aldo.vw@ufsc.br} 
           \and
          1 - Graduate Program in Computer Science, Federal University of Santa Catarina (PPGCC) - Department of Informatics and Statistics - Florianopolis, SC, Brazil \\
          2 - Image Processing and Computer Graphics Lab (LAPIX) at National Institute for Digital Convergence (INCoD)
}

\date{Received: date / Accepted: date}

\maketitle

\begin{abstract}
A challenge still to be overcome in the field of visual perception for vehicle and robotic navigation on heavily damaged and unpaved roads is the task of reliable path and obstacle detection. The vast majority of the researches have as scenario roads in good condition, from developed countries. These works cope with few situations of variation on the road surface and even fewer situations presenting surface damages. In this paper we present an approach for road detection considering variation in surface types, identifying paved and unpaved surfaces and also detecting damage and other information on other road surface that may be relevant to driving safety. We also present a new Ground Truth with image segmentation, used in our approach and that allowed us to evaluate our results. Our results show that it is possible to use passive vision for these purposes, even using images captured with low cost cameras.

\end{abstract}

\section{Introduction}
\label{sec:intro}
Visual perception for autonomous navigation has been very prominent in recent years, with a lot of papers on both path detection (\cite{rateke:2019}) and obstacle detection (\cite{rateke:2018}). Although there are excellent approaches to accomplish these two tasks, the vast majority are created based on images from developed countries, from Europe or North America, with well maintained roads, having few examples of damaged roads, nor dealing with variations in terrain type.

Both variation in surface type, possible road damages or even different surface variations (eg.: speed-bumps) are important for autonomous navigation systems, or even for an Advanced Driver-Assistance System (ADAS), because the surface conditions directly impact the way the vehicle is driven and in the comfort of the users and are also related to the vehicles conservation. For example, a pothole or a water-puddle can damage the vehicle and cause accidents. Detection of surface types, and surface variations may also be useful for Road Infrastructure Departments aiming road maintenance purposes.

A survey (\cite{cnt:2018}) from the Brazilian National Traffic Council presents a road quality evaluation where 37.0\% of the roads were classified as ``\textit{Regular}'', 9.5\% as ``\textit{Bad}'' and 4.4\% as ``\textit{Poor}''. To achieve these results, the study took into consideration the road damage as well as the lack of pavement or asphalt. It is noteworthy that this study focuses on roads under federal or state responsibility, municipal roads were not part of this survey. Another study (\cite{cabral:2018}), from East Timor, showed that 50\% of the roads are unpaved in this country.

In \cite{frisoni:2014}, an European Union (EU) study concerning road surfaces quality, it is said that compared to the human behavior while driving (eg: lack of attention, aggressively or drunk) the lack of maintenance on the road isn't the main cause of accidents, but still is one of the causes, because a pothole can damage the vehicle causing the losing of control, it is also directly related to drivers attention, as drivers reaction to variations in surface can cause them to hit other vehicles or obstacles, including pedestrians.

All these studies presents the road evaluation done by human specialists, as a mapping task not as a prediction, and they used a lot of different sensors to assist the evaluation.

The most commonly used datasets for visual perception researches are composed by images depicting good quality and well maintained roads with little variation in terrain type: The CityScapes\footnote{https://www.cityscapes-dataset.com/} (\cite{cordts:2016}) is a dataset from Germany. Also from Germany the KITTI\footnote{http://www.cvlibs.net/datasets/kitti/raw\_data.php} is a stereo images dataset (\cite{geiger:2013}). The CamVid\footnote{\footcamurl} dataset (\cite{brostow:2009}) stems from Cambridge, England. Another dataset, DIPLODOC \footnote{http://tev.fbk.eu/databases/diplodoc-road-stereo-sequence} (\cite{zanin:2013}) is from Italy and not as commonly used as the previous three.

There are new datasets that provide images of damaged roads as the RoadDamageDetector\footnote{https://github.com/sekilab/RoadDamageDetector/} dataset (\cite{maeda:2018}), from Japan. But the dataset contains only asphalt images. 
From Brazil, there is the CaRINA\footnote{http://www.lrm.icmc.usp.br/dataset} dataset (\cite{shinzato:2016}), which presents a scenario in emerging countries, depicting road damages and variations in surface type.

We also provided a dataset acquired in Brazil, the RTK\footnote{http://www.lapix.ufsc.br/pesquisas/projeto-veiculo-autonomo/datasets/?lang=en} dataset (\cite{rtk:2019}), created through low-cost cameras, including a lot of surface types variations, and damages on the road surface, even on unpaved roads.

There exist several approaches detecting potholes on the road with the use of LIDAR (Light Detection and Ranging) (\cite{kang:2017} and \cite{yu:2011}). LIDAR sensors, despite employing relatively safe laser sources, can still cause damage to the human eye (\cite{slp:2015} and \cite{ans:2005}). We understand that the pollution caused by LIDAR, which we call \textit{laser-smog}, isn't an issue yet, but in a future where smart or autonomous vehicles are a widespread reality, this could be a concern, mainly to pedestrians walking or waiting in the sidewalks near the vehicles at rush hours. 

Our goal with this work is to perform road detection with the differentiation of surface variations, in addition to  a concomitant surface damage detection. We also aim to show that it is possible to use passive vision (cameras) to detect road damages. 
It is our understanding that, in dealing with common problems found on roads in emerging countries, but which can also occur in developed countries and are of utmost importance to vehicle behavior, both for the sake of vehicle conservation and especially for safety, we are advancing the state-of-the art of path detection.

The remaining sections of this paper are organized as follows: Section \ref{sec:relwork} shows some related works that dealt with road detection or road damage detection using passive vision. In Section \ref{sec:matmet} we provide a brief description about our dataset and our setup, hardware and software, used. Our approach is presented In Section \ref{sec:ourApp}, followed by the results in Section \ref{sec:results}. Finally, in Section \ref{sec:cdfw} we conclude this paper summarizing the information, with a discussion about our results, besides listing some topics for future works.

\section{Related work}
\label{sec:relwork}
In a previous work, we performed a Systematic Literature Review (SLR) on Road Detection research that employed Passive Computer Vision (\cite{rateke:2019}). In this SLR, although we found several papers, we could not identify any work that dealt with road damage detection or road features other than painted road markings (\cite{ardiyanto:2017}, \cite{jia:2017}, \cite{yuan:2015}, \cite{zu:2015} and \cite{shi:20162}). Still in the road detection SLR, it is noticed that only 32.1\% works did the detection in different surface types (eg: \cite{li:2016}, \cite{wang:2016}, \cite{nguyen:2017} and \cite{valente:2017}). Only 4 showed results with path detection working during the transition between surface types (\cite{guo:2011}, \cite{guo:2012}, \cite{ososinski:2012} and \cite{cristoforis:2016}), in situations where transitions were not too different, such as focus between asphalt to paved. These approaches did not differentiate the type of surface, i.e. regardless of whether it is asphalt, unpaved or paved, everything was considered as road. Also, none of the approaches was aimed towards detecting other road features, even if there were approaches aimed at path detection on unpaved roads (eg: \cite{wang2:2015} and \cite{xiao:2015}).

In another SLR for Road Obstacle Detection (\cite{rateke:2018}), we identified a few papers dealing specifically with potholes on the road, using stereo vision techniques, where potholes were defined as a \textit{negative obstacles} (\cite{herghelegiu:2017}, \cite{karunasekera:2017}). However, these works deal only with pothole detection, depending on the depth information of the scene and without other information or features from the road.

In both SLRs it was possible to notice the recent increase of the use of Convolutional Neural Networks (CNN) applications in the tasks of visual perception for navigation. A recent work (\cite{maeda:2018}), which employs Deep Neural Networks, detects and classifies asphalt damages with Bounding Box markings, but does not deal with surface types variation or damages in other surfaces types.

The authors from a paper where they describe path detection as a guidance for blind people affirm they deal with potholes and water-puddles, but the paper did not present these results (\cite{soveny:2015}). Other two papers (\cite{rankin:2008} and \cite{rankin:2010}) deal with larger water-puddle detection in off-road scenarios for navigation purposes with the use of stereo vision methods. They do not deal with other road features, not even the road detection itself.

There are other works that perform the pothole detection, but differ from our scope, because they are with images very close and vertical to the pothole, serving as a mapping task and not for prediction (eg.: \cite{eriksson:2008}, \cite{koch:2011}, \cite{huidrom:2013}, \cite{tedeschi:2017} and \cite{banharnsakun:2017}). Our goal is to identify surface features before the vehicle passes. These approaches also did not dealt with path detection and other path features.

\section{Material and methods}
\label{sec:matmet}
In this section we present the dataset we have used in our experiments, the RTK dataset (\cite{rtk:2019}) and the corresponding Ground Truth (GT) that we created to train and validate our experiments, in Subsection \ref{sec:rtkds}. In Subsection \ref{sec:ourSet} we list the hardware and software configurations relevant to our experiment.

\subsection{RTK dataset}
\label{sec:rtkds}
The RTK dataset is composed by 77547 images captured by a low-cost camera with low-resolution, which could increase the challenge of our application. This dataset was primarily used for a surface type and quality classification application (\cite{rtk:2019}). The dataset is composed of images captured during the daytime, containing lighting variations and plenty of shadows on the road. Contains images of asphalt roads, unpaved, different paved roads and transitions between surface types. Also contains variations in the quality of these roads, with potholes, water-puddles, speed-bumps, patches, etc. 

\begin{figure}[ht]\centering
	\includegraphics[width=\linewidth]{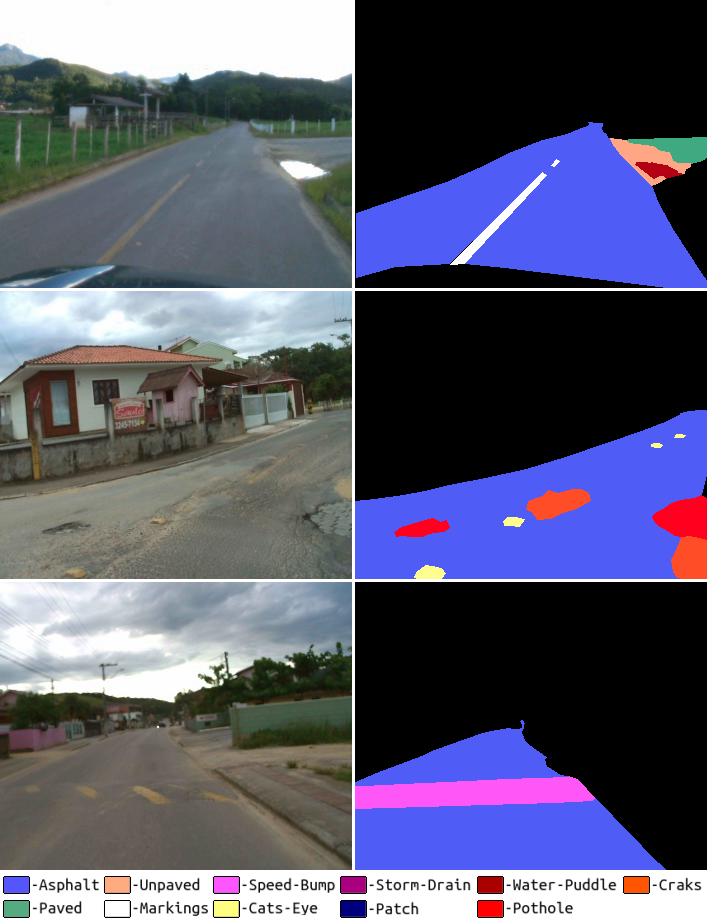}
	\caption{Samples images from RTK dataset and our segmented GT}
	\label{fig:sampleGT01}
\end{figure}

\begin{figure}[ht]\centering
	\includegraphics[width=\linewidth]{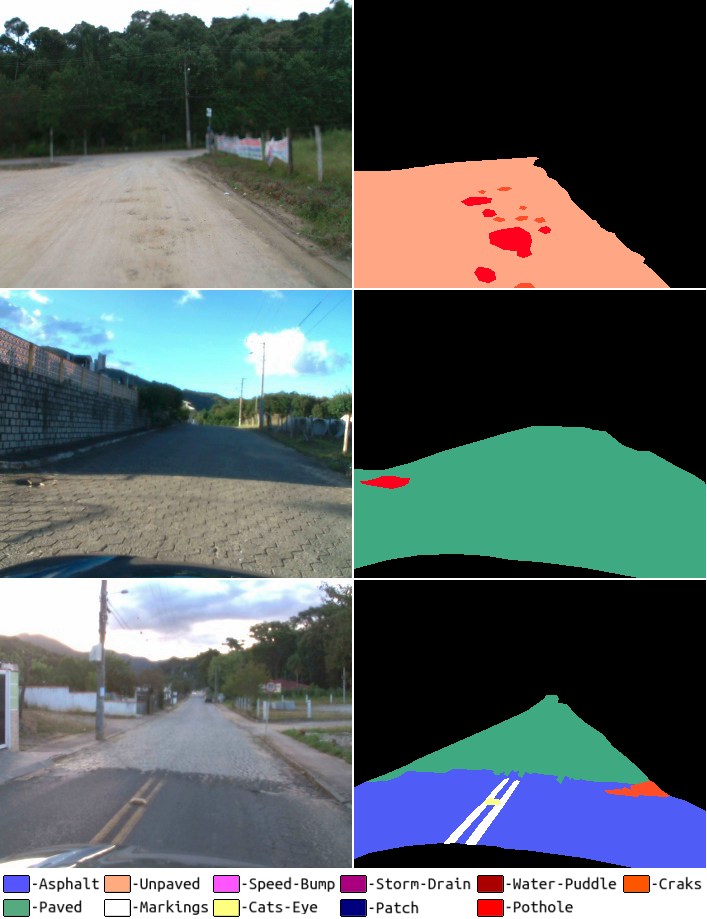}
	\caption{Samples images from RTK dataset and our segmented GT}
	\label{fig:sampleGT02}
\end{figure}

Based on RTK dataset, we create a segmented GT for our experiments. We took 701 images with different situations for label annotation (Figure \ref{fig:sampleGT01} and Figure \ref{fig:sampleGT02}). In our GT we defined the following classes:

\begin{itemize}
    \item \textbf{Background}, everything being unrelated to the road surface;
    \item \textbf{Asphalt}, roads with asphalt surface;
    \item \textbf{Paved}, different pavements (eg.: Cobblestone);
    \item \textbf{Unpaved}, for unpaved roads;
    \item \textbf{Markings}, to the road markings;
    \item \textbf{Speed-Bump}, for the speed-bumps on the road;
    \item \textbf{Cats-Eye}, for the cats-eye found on the road, both on the side and in the center of the path;
    \item \textbf{Storm-Drain}, usually at the side edges of the road;
    \item \textbf{Patch}, for the various patches found on asphalt road;
    \item \textbf{Water-Puddle}, We use this class also for muddy regions;
    \item \textbf{Pothole}, for different types and sizes of potholes, no matter if they are on asphalt, paved or unpaved roads;
    \item \textbf{Cracks}, Used in different road damages, like ruptures.
\end{itemize}

\subsection{Our Setup}
\label{sec:ourSet}
We did our experiments on Google Colaboratory (Google Colab), a cloud service based on Jupyter notebooks, which is an interactive web-based environment for document creation. Google Colab also offers free GPU. In our experiments we can use Tesla K80 GPU with 12GB memory and Tesla P100-PCIE GPU with 16GB memory. We also made use of fast.ai library (\cite{howard:2018}), an open source library for deep learning, based on PyTorch.

\section{Our approach}
\label{sec:ourApp}
Our approach consists in the use of deep learning for road surface semantic segmentation, considering variations in road surface and through low resolution images. That is, label each pixel of the image as the corresponding class as we defined in our GT. To do this, we need to train a Convolutional Neural Network (CNN), and find the best possible configuration that contains reasonable accuracy for all classes.

In our experiments we used the U-NET (\cite{ronneberger:2015}) for semantic segmentation, which is a convolutional network architecture designed to accomplish the task of fast semantic segmentation in medical imaging, but successfully applied to many other approaches. This architecture has two main parts, one being the encoder, used to extract features from the image, with a traditional CNN structure (including convolution layers and max-pooling layers). The encoder starts with input size image and makes these inputs small. The other part, the decoder, symmetrical at the encoder, makes the process of increasing back to the original size. U-NET accepts input images of any size.

The fast.ai library has different pre-trained models, we did our experiments with resnet34 and resnet50, resnet34 being faster to train and requiring less memory. ResNet are residual CNN models, with skip-connections, allowing sections to be skipped. Each Residual Block has two connections, one connection skipping over that series of convolutions and functions and the other connection going through layers without skipping (\cite{he:2016}). This helps maintain important features of the early layers. Resnet is used on the U-NET encoder part, and the fast.ai library will automatically build the symmetrical decoder part of the U-NET.

As data augmentation we used only randomly horizontal rotations and the perspective warping, which is by default in fast.ai library, and we think its the only one necessary and that makes sense in road detection scenario. The data augmentation allows the increase of training images, because besides the original images, also uses the images with the transformations (horizontal rotation and wrap). The fast.ai library also has an option to make the same data augmentation into the original and into the respective mask images.

In our early experiments we realized that the network could reasonably identify the asphalt, paved and unpaved pixels beyond the background, but the other classes resulted in very low accuracy. Due to this disparity, we checked in our GT the number of segmented pixels in each class and found what was already visually noticeable, that there is a great imbalance between the classes. The ratio of each class's pixels to the total pixels in GT is as follows:

\begin{itemize}
    \item \textbf{Background} = 65.86\%;
    \item \textbf{Asphalt} = 12.90\%;
    \item \textbf{Paved} = 10.50\%;
    \item \textbf{Unpaved} = 9.22\%;
    \item \textbf{Marking} = 0.78\%;
    \item \textbf{Speed-Bump} = 0.06\%;
    \item \textbf{Cats-Eye} = 0.02\%;
    \item \textbf{Storm-Drain} = 0.02\%;
    \item \textbf{Patches} = 0.22\%;
    \item \textbf{Water-Puddle} = 0.03\%;
    \item \textbf{Pothole} = 0.06\%;
    \item \textbf{Cracks} = 0.33\%.
\end{itemize}

This is a situation that, for example, if we collect more images with potholes, we will also collect even more background pixels or some of the road surface types, maintaining the disparity. Then we added weights to each class to simulate that they all had a similar proportion of pixels in the training step. With the weights the accuracy of the smaller classes has considerably improved, but with the burden of losing accuracy in the asphalt, paved and unpaved classes.

\begin{figure}[ht]\centering
	\includegraphics[width=\linewidth]{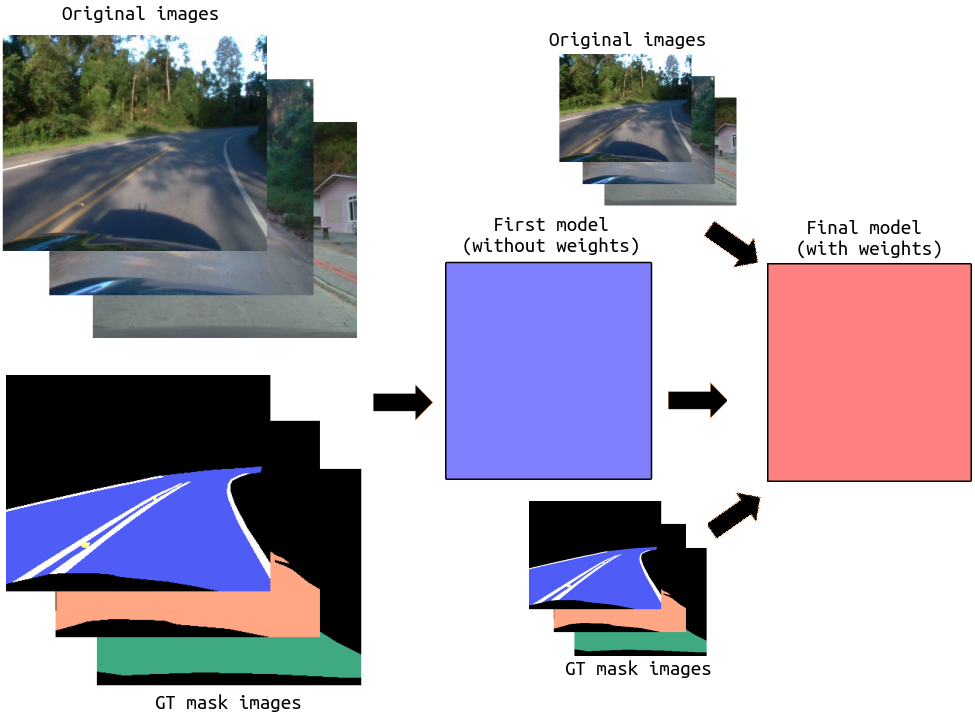}
	\caption{Best results approach}
	\label{fig:steps}
\end{figure}

Trying to get more accurate values, after different experiments with different configurations (presented in Section \ref{sec:results}) we came up with a solution. First we train a model without the use of weights, in this model the network identifies well the surface patterns but does not have good accuracy for the smaller classes. We then use the previously trained model as the basis for the next model, with the weights in the classes (Figure \ref{fig:steps}). This configuration generated the most satisfactory and yet balanced accuracy results.

\section{Results}
\label{sec:results}
As we said in Section \ref{sec:ourApp} we did experiments with different configurations. Starting from smaller datasets to bigger ones, also with no changes at all, and using weights too. In order to make this comparison we tried every model in each configuration during 25 epochs. The configurations we tried are listed as follows:

\begin{itemize}
    \item \textbf{r34-S}: ResNet34. Only one model. Without weight;
    \item \textbf{r34-SW}: ResNet34. Only one model. With weight;
    \item \textbf{r34-I}: ResNet34. Three models, smaller to increasing dataset. First model with images divided by 4, second model with images divided by 2 and third with original sizes. Without weight;
    \item \textbf{r34-IW}: ResNet34. Three models, smaller to increasing dataset. First model with images divided by 4, second model with images divided by 2 and third with original sizes. With weight;
    \item \textbf{r34-DW}: ResNet34. Two models. First model without weight and second model with weights;
    \item \textbf{r50-S}: ResNet50. Only one model. Without weight;
    \item \textbf{r50-SW}: ResNet50. Only one model. With weight;
    \item \textbf{r50-I}: ResNet50. Three models, smaller to increasing dataset. First model with images divided by 4, second model with images divided by 2 and third with original sizes. Without weight;
    \item \textbf{r50-IW}: ResNet50. Three models, smaller to increasing dataset. First model with images divided by 4, second model with images divided by 2 and third with original sizes. With weight;
    \item \textbf{r50-DW}: ResNet50. Two models. First model without weight and second model with weights.
\end{itemize}

The values obtained on each setting are shown in Table \ref{tbl:expres}. Based on these results, we note that networks with only one trained model and weightless (\textbf{r34-S} and \textbf{r50-S}), despite having a good total accuracy, also have the worst results for the smaller classes. The same is true for networks with 3 models with increasing image size (\textbf{r34-I} and \textbf{r50-I}).

The previous models (one model and three models with increasing image size) with the classes weight adjustment, despite having a loss in total accuracy, showed a great improvement in the accuracy of the smaller classes. However they also showed considerable loss in the values of the larger classes, the road surfaces classes (\textbf{r34-SW}, \textbf{r34-IW}, \textbf{r50-SW} and \textbf{r50-IW}).

Both approaches using an initial model with no class weights, followed by the final model with weight in the classes (\textbf{r34-DW} and \textbf{r50-DW}) showed good results for the smaller classes without having a large loss in road surface classes and also maintaining a good total accuracy value.

\begin{sidewaystable}[]
\centering
\caption{Results from different settings}
\label{tbl:expres}
\begin{adjustbox}{width=\columnwidth}
\begin{tabular}{|c|c|c|c|c|c|c|c|c|c|c|c|c|c|}
\hline
\rowcolor[HTML]{C0C0C0} 
 & \textbf{Background} & \textbf{Asphalt} & \textbf{Paved} & \textbf{Unpaved} & \textbf{Markings} & \textbf{Speed-Bump} & \textbf{Cats-Eye} & \textbf{Storm-Drain} & \textbf{Patchs} & \textbf{Water-Puddle} & \textbf{Pothole} & \textbf{Cracks} & \textbf{Total} \\ \hline
\cellcolor[HTML]{C0C0C0}\textbf{r34-S} & 98,00\% & 93,00\% & 88,00\% & 74,00\% & 43,00\% & 0,00\% & 0,00\% & 0,00\% & 0,01\% & 0,00\% & 0,00\% & 0,07\% & \cellcolor[HTML]{EFEFEF}\textbf{94,00\%} \\ \hline
\rowcolor[HTML]{FFFFC7} 
\cellcolor[HTML]{C0C0C0}\textbf{r34-SW} & 88,00\% & 71,00\% & 76,00\% & 61,00\% & 71,00\% & 11,00\% & 94,00\% & 98,00\% & 69,00\% & 71,00\% & 66,00\% & 45,00\% & \cellcolor[HTML]{FFFC9E}\textbf{89,00\%} \\ \hline
\cellcolor[HTML]{C0C0C0}\textbf{r34-I} & 98,00\% & 89,00\% & 86,00\% & 74,00\% & 50,00\% & 0,00\% & 0,00\% & 0,00\% & 11,00\% & 0,00\% & 0,00\% & 0,47\% & \cellcolor[HTML]{EFEFEF}\textbf{95,00\%} \\ \hline
\rowcolor[HTML]{FFFFC7} 
\cellcolor[HTML]{C0C0C0}\textbf{r34-IW} & 84,00\% & 57,00\% & 67,00\% & 62,00\% & 68,00\% & 75,00\% & 79,00\% & 95,00\% & 62,00\% & 60,00\% & 72,00\% & 38,00\% & \cellcolor[HTML]{FFFC9E}\textbf{83,00\%} \\ \hline
\cellcolor[HTML]{C0C0C0}\textbf{r34-DW} & 92,00\% & 85,00\% & 87,00\% & 79,00\% & 73,00\% & 58,00\% & 86,00\% & 86,00\% & 78,00\% & 89,00\% & 66,00\% & 51,00\% & \cellcolor[HTML]{EFEFEF}\textbf{92,00\%} \\ \hline
\rowcolor[HTML]{FFFFC7} 
\cellcolor[HTML]{C0C0C0}\textbf{r50-S} & 97,00\% & 88,00\% & 73,00\% & 72,00\% & 19,00\% & 0,00\% & 0,00\% & 0,00\% & 0,00\% & 0,00\% & 0,00\% & 0,00\% & \cellcolor[HTML]{FFFC9E}\textbf{87,00\%} \\ \hline
\cellcolor[HTML]{C0C0C0}\textbf{r50-SW} & 91,00\% & 78,00\% & 83,00\% & 70,00\% & 71,00\% & 58,00\% & 85,00\% & 95,00\% & 74,00\% & 95,00\% & 59,00\% & 45,00\% & \cellcolor[HTML]{EFEFEF}\textbf{90,00\%} \\ \hline
\rowcolor[HTML]{FFFFC7} 
\cellcolor[HTML]{C0C0C0}\textbf{r50-I} & 98,00\% & 94,00\% & 89,00\% & 76,00\% & 67,00\% & 14,00\% & 18,00\% & 20,00\% & 38,00\% & 5,03\% & 20,00\% & 13,00\% & \cellcolor[HTML]{FFFC9E}\textbf{96,00\%} \\ \hline
\cellcolor[HTML]{C0C0C0}\textbf{r50-IW} & 87,00\% & 73,00\% & 72,00\% & 67,00\% & 79,00\% & 85,00\% & 82,00\% & 88,00\% & 70,00\% & 73,00\% & 69,00\% & 49,00\% & \cellcolor[HTML]{EFEFEF}\textbf{87,00\%} \\ \hline
\rowcolor[HTML]{FFFFC7} 
\cellcolor[HTML]{C0C0C0}\textbf{r50-DW} & 90,00\% & 80,00\% & 79,00\% & 76,00\% & 72,00\% & 93,00\% & 93,00\% & 94,00\% & 75,00\% & 97,00\% & 81,00\% & 48,00\% & \cellcolor[HTML]{FFFC9E}\textbf{93,00\%} \\ \hline
\end{tabular}%
\end{adjustbox}
\end{sidewaystable}

Continuing and aiming to check if training with more epochs we can get better results. We chose between the two best performing approaches the \textbf{r34-DW}, which showed great results in the experiments and we made a new training, and being with resnet34 requires less of the GPU and we were able to train with a larger batch size and have a faster result. We then trained with 100 epochs for each model from this approach, the first model without weights and then the second using weights. The results obtained are presented in Table \ref{tbl:finalr34}.

\begin{table}[]
\centering
\caption{Final accuracy results. \textbf{r34-DW} during 100 epochs.}
\label{tbl:finalr34}
\begin{tabular}{lll}
\hline\noalign{\smallskip}
label & accuracy \\
\noalign{\smallskip}\hline\noalign{\smallskip}
Background & 97\% \\
Asphalt & 92\% \\
Paved & 94\% \\
Unpaved & 90\% \\
Marking & 93\% \\
Speed-Bump & 69\% \\
Cats-Eye & 94\% \\
Storm-Drain & 97\% \\
Patches & 97\% \\
Water-Puddle & 90\% \\
Pothole & 97\% \\
Cracks & 72\% \\
\noalign{\smallskip}\hline\noalign{\smallskip}
Total & 97\% \\
\noalign{\smallskip}\hline
\end{tabular}
\end{table}

\begin{figure}[htb]\centering
	\includegraphics[width=\linewidth]{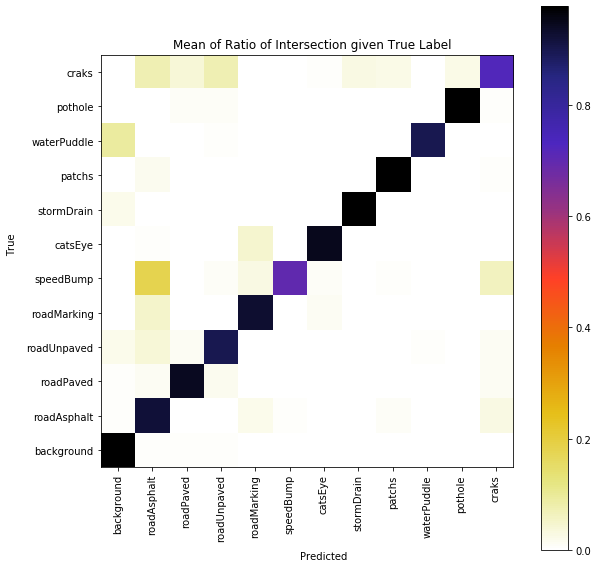}
	\caption{Confusion matrix for \textbf{r34-DW}}
	\label{fig:confmatrs34}
\end{figure}

In the confusion matrix presented in Figure \ref{fig:confmatrs34} we can analyze the results from Table \ref{tbl:finalr34} And find out where the major pixel labeling errors occurred. The \textit{Cracks} class, for example, which had 72\% accuracy, had its biggest confusions with the road surface classes. It also presents errors as being in the \textit{Patches}, \textit{Storm-Drain} and \textit{Pothole} classes. The \textit{Speed-Bump} class, which ended up being the least accurate, had much of the error as \textit{Asphalt} class, and slightly less as \textit{Cracks} and \textit{Marking}. The \textit{Water-puddle} class that had a hit index of 90\% concentrated the errors that occur as being \textit{Background}. The \textit{Marking} class has the biggest confusion related to \textit{Asphalt}. The \textit{Unpaved} class has errors as being \textit{Asphalt}.

\begin{figure*}[]\centering
	\includegraphics[width=\textwidth,height=\textheight,keepaspectratio]{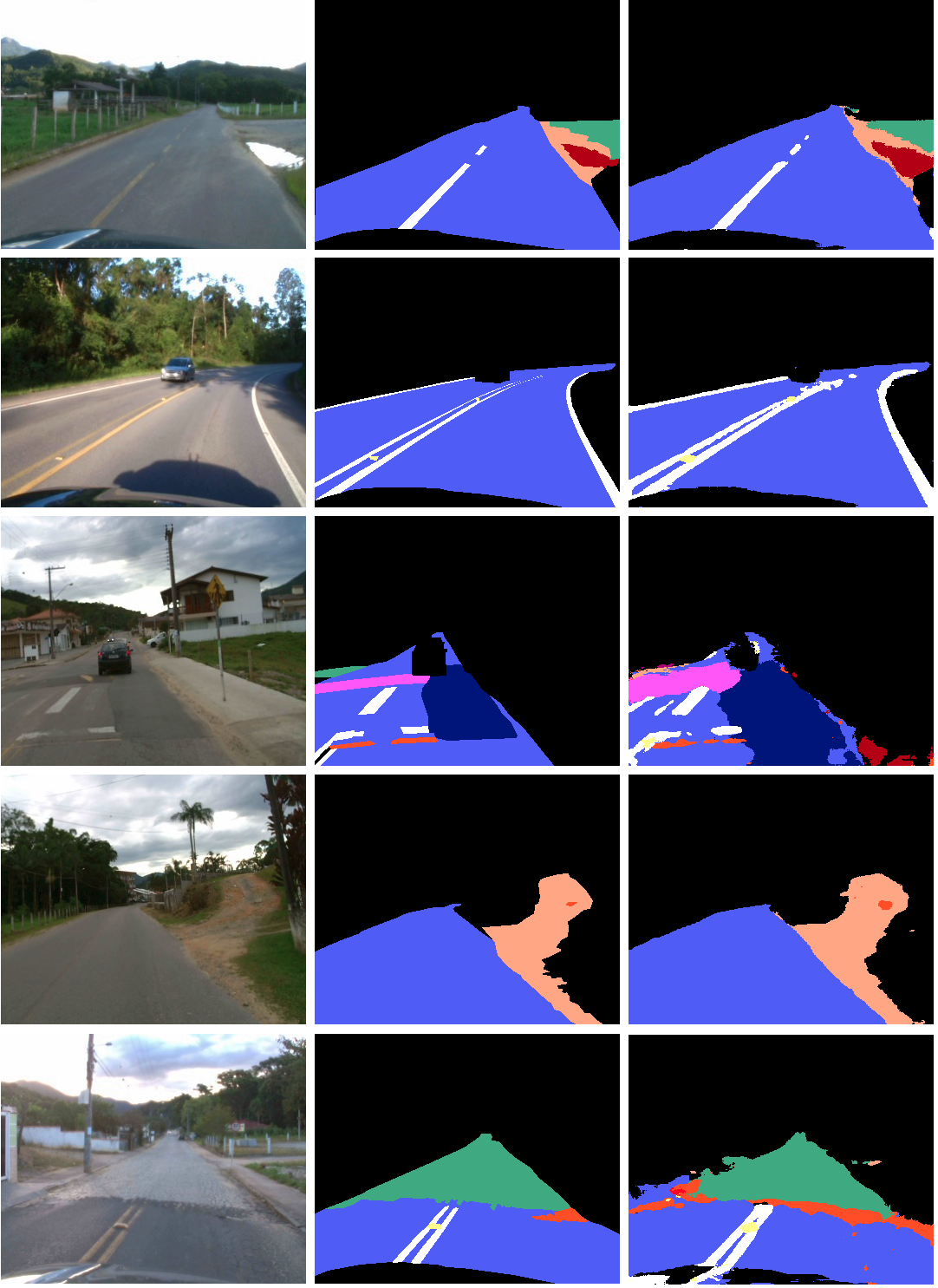}
	\caption{Results in validation dataset. Left: original. Middle: GT mask. Right: Result}
	\label{fig:finalRes}
\end{figure*}

\begin{figure*}[]\centering
	\includegraphics[width=\textwidth,height=\textheight,keepaspectratio]{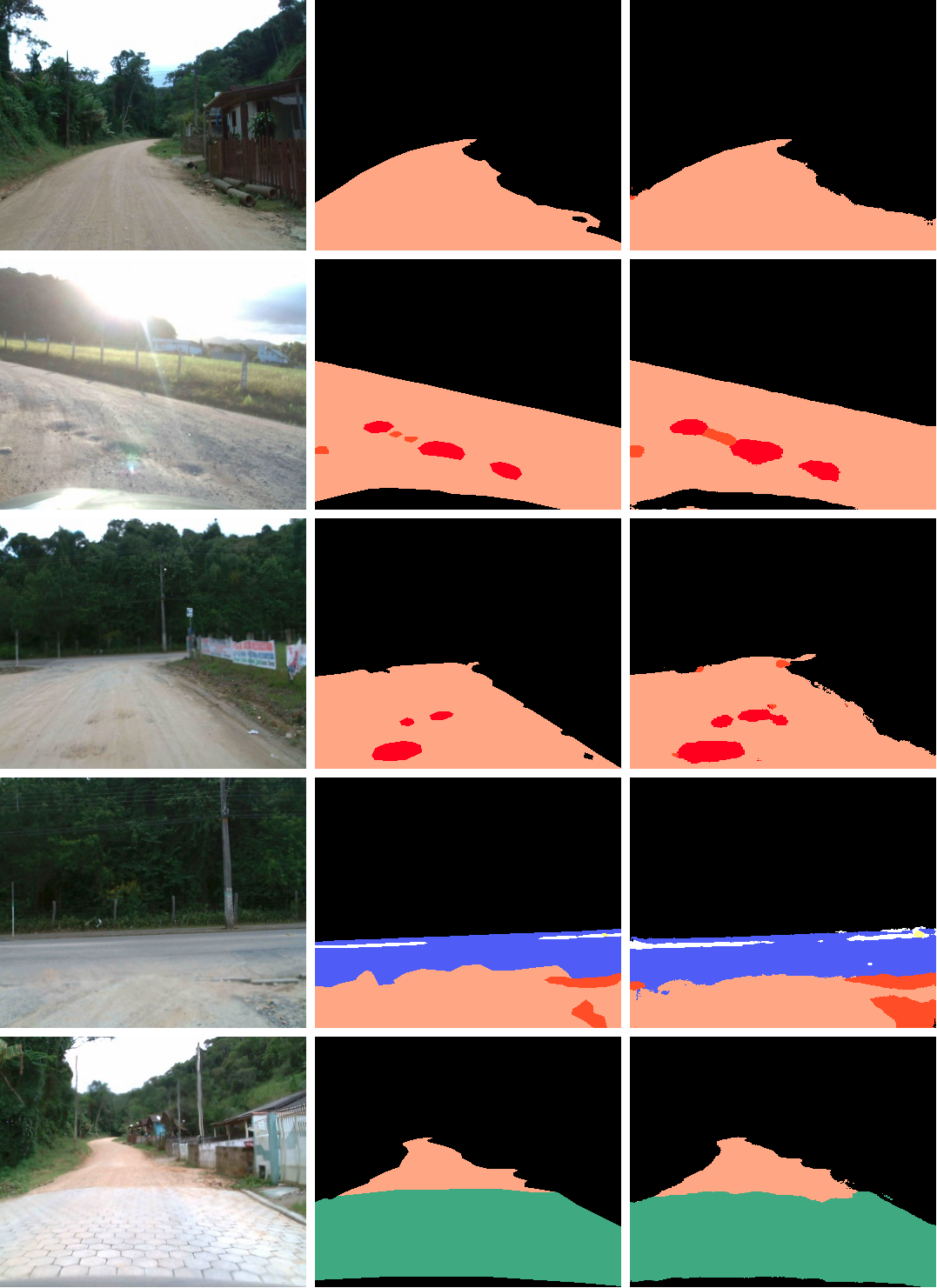}
	\caption{Results in validation dataset. Left: original. Middle: GT mask. Right: Result}
	\label{fig:finalRes02}
\end{figure*}

\begin{figure*}[]\centering
	\includegraphics[width=\textwidth,height=\textheight,keepaspectratio]{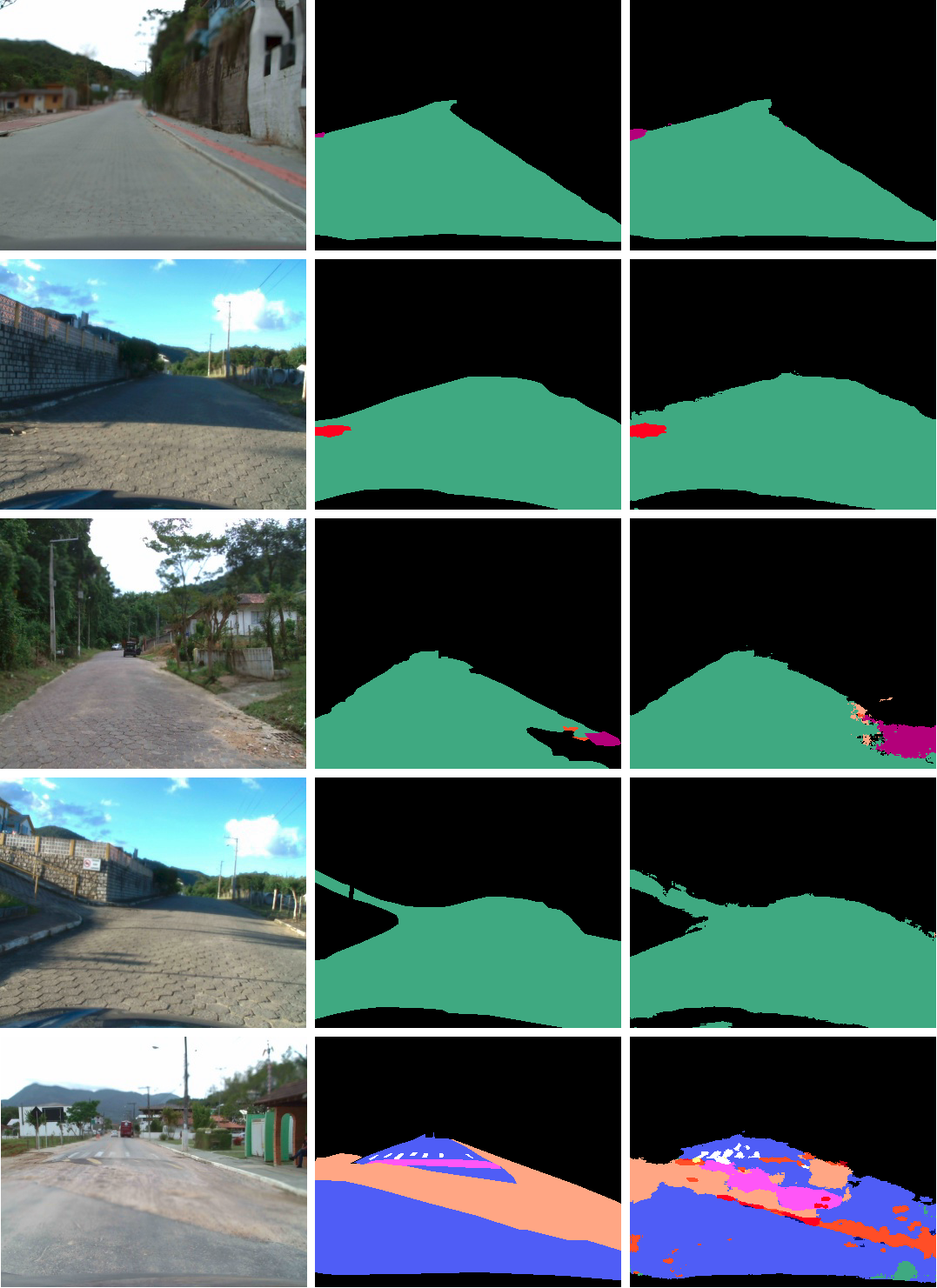}
	\caption{Results in validation dataset. Left: original. Middle: GT mask. Right: Result}
	\label{fig:finalRes03}
\end{figure*}

We also present some prediction results from the dataset validation images in the Figure \ref{fig:finalRes}, Figure \ref{fig:finalRes02} and Figure \ref{fig:finalRes03}. Comparing the original images (left column), the GT mask images (middle column) and the prediction images (right column). In Figure \ref{fig:finalRes} the first, second and fourth rows show an almost exact result compared to GT. The third and the fifth rows show visible discrepancies from the GT, but still very close to the correct one.

In Figure \ref{fig:finalRes02} first and last rows show a very accurate result. While the other rows show slight variations when comparing the results with the GT images. Finally, in Figure \ref{fig:finalRes03}, The first four rows show results very close to the GT, although with some variations. However, the last row, presents a greatly deforming. Still, it indicates that exists a speed-bump near in front, as well as variations in the surface.

\section{Discussion, Conclusions and Future Work}
\label{sec:cdfw}
Despite the great advances on the state-of-the art in tasks related to the visual perception for navigation, especially in recent years, considering the advancement of CNN architectures, where applications are beginning to present increasingly accurate and reliable results. We believe that there are still many challenges ahead, especially if we consider the issue of road surface quality and variation.

Although being a more prevalent issue in emerging countries, identifying surface conditions and features is important in any scenario, because it influences how the vehicle should behave or how it should be driven, enabling safer decision making. Surface features detection can be useful for Autonomous Vehicle navigation, for Advanced Driver-Assistance Systems (ADAS), and even for Road Maintenance Departments.

We present in this paper a new GT for road surface features semantic segmentation using low resolution images captured with low cost cameras. We also present the application of deep learning using this GT for the surface features semantic segmentation. We compare different settings and also present the validation values, using the chosen setting, in the results section.

We obtained good results with the configuration based upon first training a model without weights in the classes, and using this model as pre-training for the next model, where we balanced the dataset with weights in the classes, aiming to maintain a good accuracy value for the smaller classes, especially for pothole and water-puddle, without losing much accuracy in the larger surface classes (asphalt, paved and unpaved).

\subsection{Future Work}
\label{sec:fw}
In this work we raised some possibilities and questions while performing our experiments. One question that may result in further experimentation is whether to differentiate more some label categories, instead of employing them always in the same, generalized way, regardless of road surface. For example, we define all pavement damages as \textit{Cracks}, regardless of whether they occur on asphalt, paved or unpaved roads. The types of damage, however, may have different features on each surface type and perhaps differentiating them may improve the accuracy of the \textit{Cracks} class, the second lowest accuracy result. In order to improve results, it would be possible to try a finer definition of the general category of \textit{cracklike damages}: \textit{Asphalt Cracks}, \textit{Paved Cracks} or \textit{Unpaved Cracks}, thus enabling a better differentiation. The same idea goes for the other classes of our dataset. 

\begin{figure}[ht]\centering
	\includegraphics[width=\linewidth]{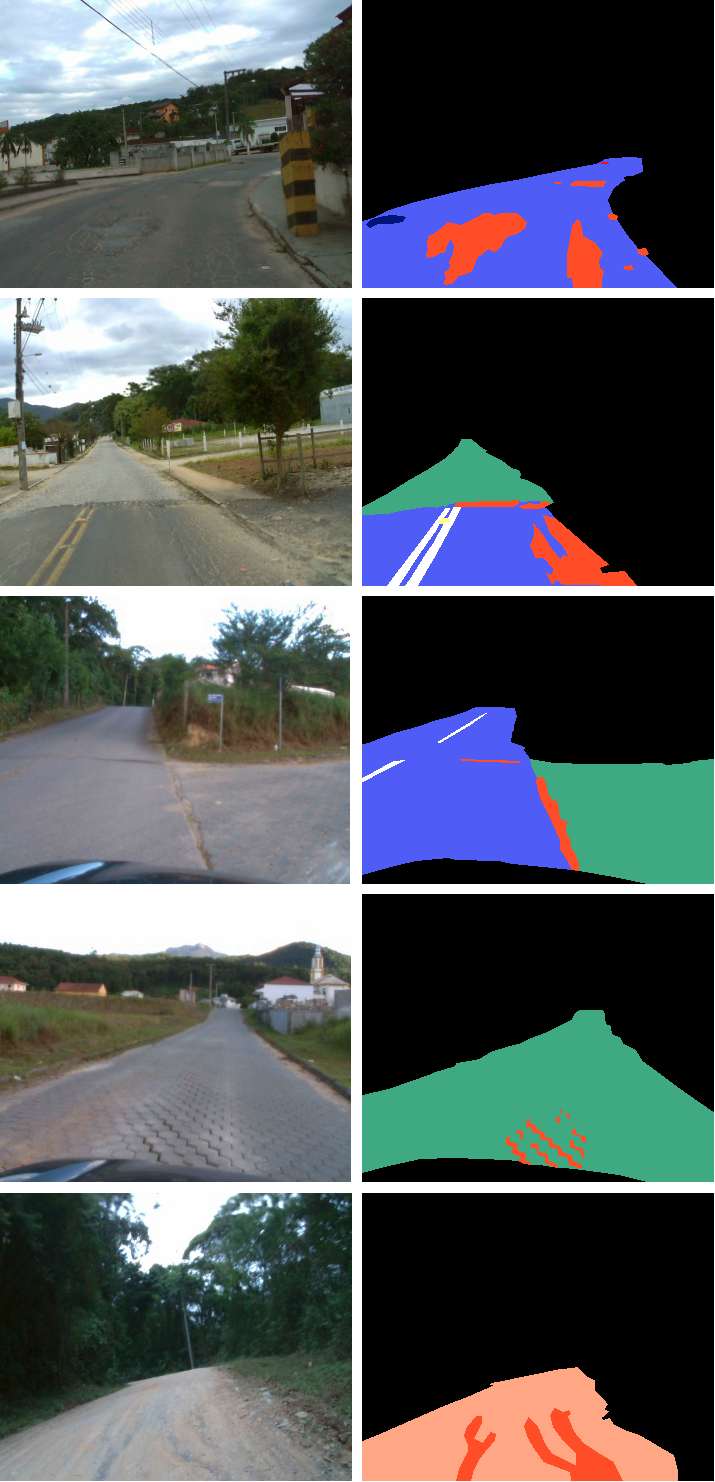}
	\caption{Samples images from RTK dataset. Original (left). GT (right). Which the \textit{Crack} class in situations that could be in new specific classes.}
	\label{fig:newClassCracks}
\end{figure}

Another approach, also focusing more on the \textit{Cracks} category is not only to separate damages by surface types, but also to create new, more specific classes, as we already did to the \textit{Pothole} and \textit{Water-Puddle} classes. In Figure \ref{fig:newClassCracks} we show some situations where the \textit{Crack} class could be separated in new and more specific classes.

Concluding, based on the results obtained here, we determined that using only standard resolution monocular video streams we were able to reliably extract useful information on the road surface status. This information could, e.g., be used by an intelligent system, to determine threats such as the distance and position of potholes, water-puddles and other damages and obstacles. Finally, we show that there are still other challenges, such as identifying surface type and surface variations on a rainy day, on a foggy environment or even at night.


%
\section*{Conflicts of interest}
The authors declare that there are no conflicts of interest.

\bibliographystyle{spbasic}      
\bibliography{bibliography}   

\end{document}